\documentclass[conference]{IEEEtran}
\IEEEoverridecommandlockouts
\usepackage{cite}
\usepackage{amsmath,amssymb,amsfonts}
\usepackage{algorithmic}
\usepackage{balance}
\usepackage{graphicx}
\usepackage{textcomp}
\usepackage{tabularx}
\usepackage{xcolor}
\usepackage{multirow}
\usepackage{subfigure}
\def\BibTeX{{\rm B\kern-.05em{\sc i\kern-.025em b}\kern-.08em
    T\kern-.1667em\lower.7ex\hbox{E}\kern-.125emX}}

\title{Practical Battery Health Monitoring using Uncertainty-Aware Bayesian Neural Network\\
\thanks{This work was supported in part by A*STAR under its MTC Programmatic (Award M23L9b0052), MTC Individual Research Grants (IRG) (Award M23M6c0113), the Ministry of Education, Singapore, under the Academic Research Tier 1 Grant (Grant ID: GMS 693), SIT’s Ignition Grant (STEM) (Grant ID: IG (S) 2/2023 – 792), and Future Communications Research \& Development Programme (FCP) under Grant FCP-SIT-TG-2022-007.}
}

\makeatletter
\newcommand{\linebreakand}{%
  \end{@IEEEauthorhalign}
  \hfill\mbox{}\par
  \mbox{}\hfill\begin{@IEEEauthorhalign}
}
\makeatother

\author{\IEEEauthorblockN{Yunyi Zhao, Wei Zhang$^*$\thanks{$^*$ Corresponding author. Email: wei.zhang@singaporetech.edu.sg.}}
\IEEEauthorblockA{
\textit{Singapore Institute of Technology}
\\\{yunyi.zhao, wei.zhang\}@singaporetech.edu.sg
}
\and
\IEEEauthorblockN{Qingyu Yan}
\IEEEauthorblockA{
\textit{Nanyang Technological University, Singapore}
\\alexyan@ntu.edu.sg
}
\linebreakand 
\IEEEauthorblockN{Man-Fai Ng}
\IEEEauthorblockA{
\textit{Agency for Science, Technology and
Research}
\\ngmf@ihpc.a-star.edu.sg
}
\and
\IEEEauthorblockN{B. Sivaneasan}
\IEEEauthorblockA{
\textit{Singapore Institute of Technology}
\\sivaneasan@singaporetech.edu.sg
}
\and
\IEEEauthorblockN{Cheng Xiang}
\IEEEauthorblockA{
\textit{National University of Singapore}
\\elexc@nus.edu.sg
}
}

\begin{document}
\bstctlcite{IEEEexample:BSTcontrol}

\maketitle

\begin{abstract}
Battery health monitoring and prediction are critically important in the era of electric mobility with a huge impact on safety, sustainability, and economic aspects. Existing research often focuses on prediction accuracy but tends to neglect practical factors that may hinder the technology's deployment in real-world applications. In this paper, we address these practical considerations and develop models based on the Bayesian neural network for predicting battery end-of-life. Our models use sensor data related to battery health and apply distributions, rather than single-point, for each parameter of the models. This allows the models to capture the inherent randomness and uncertainty of battery health, which leads to not only accurate predictions but also quantifiable uncertainty. We conducted an experimental study and demonstrated the effectiveness of our proposed models, with a prediction error rate averaging 13.9\%, and as low as 2.9\% for certain tested batteries. Additionally, all predictions include quantifiable certainty, which improved by 66\% from the initial to the mid-life stage of the battery. This research has practical values for battery technologies and contributes to accelerating the technology adoption in the industry. 
\end{abstract}


\section{Introduction}
Electric mobility is increasingly recognized as the future of transportation, largely driven by the need for sustainability and reduced greenhouse gas emissions. Electric vehicles (EVs) currently make up about 15\% of new car sales globally, and the market share is anticipated to rise to 35\% by 2030, according to the International Energy Agency. In countries like China, the percentage could be even higher, potentially reaching 40\%. Central to the EV ecosystem is battery, a critical component that determines not only EV's mobility performance such as range, but also its impact on sustainability and emission. Battery therefore has become a research topic for researchers across various fields, e.g., material science and computer science, and attracted significant investment worldwide. 

Monitoring battery health is a critical aspect of battery technologies. A degraded battery poses significant safety risks and potentially can lead to hazardous situations such as fires and explosions. Accurate health monitoring can provide early warnings to ensure the safety of EV owners and passengers. Besides, the monitoring helps access the battery's operational condition and estimate its end-of-life (EoL). As the battery approaches its EoL, it can be recycled or repurposed \cite{zhao2024batsort} for less demanding applications like power grid secondary energy storage. This is especially beneficial given the high cost of batteries, which can exceed ten thousand dollars for a typical capacity of 100kWh. 

Existing works for battery health monitoring can be broadly categorized into several groups. One group utilizes traditional methods such as electrical engineering and simulations. This includes practices like monitoring voltage and current or employing multi-physics simulations to infer battery health. Another group follows data-driven and machine learning (ML) strategies. Its initial effort is time-series-based forecasting where various ML forecasting algorithms have been explored and tested. \cite{zhao2022lithium} proposed a fusion neural network (NN) model combining a broad learning system and long short-term memory (LSTM), and \cite{9208399} developed an LSTM model integrated with incremental capacity analysis for predicting battery capacity in the next cycle. Additionally, some studies argue that the above methods do not fully utilize the data available and extract features from the raw data to better correlate with battery health. For instance, \cite{severson2019data} demonstrated the effectiveness of the features derived from degradation voltage curves, while \cite{alipour2022improved,fei2021early} examined feature selection schemes. These features are further integrated with complex ML models to enhance prediction accuracy. For example, \cite{8418374} introduced an auto-encoder approach, and \cite{shen2020hybrid} designed a relevance vector machine combined with convolutional NN achieving merely 12\% of EoL prediction error. However, the primary emphasis of these works remains on prediction accuracy, often at the expense of broader and practical insights into other critical aspects like uncertainty. Some works like \cite{9694435} took uncertainty in battery health prediction, but the focus is the next cycle prediction where uncertainty is not an urgent concern.

In this paper, we present practical solutions for predicting the EoL of batteries by integrating uncertainty into our models and dynamically updating the predictions at various stages of battery usage. We provide battery owners detailed information with not only the expected EoL but also a probability distribution that indicates potential earlier or later EoL occurrences in practical scenarios. This approach brings various benefits, e.g., enabling preparations for EoL earlier than expected when batteries begin to pose significant risks. Given more and more battery health information over the course of the battery's usage, our predictions will be refined and we expect more accurate EoL estimations and increased certainty.

Specifically, we adopt a Bayesian neural network (\textsc{Bnn}), an ML algorithm within the NN family. \textsc{Bnn} is known for its capability of quantifying prediction uncertainty. Our approach involves a sensor network that measures and monitors various aspects of battery health, e.g., discharge capacity and temperature. We process the raw sensor data to extract features for \textsc{Bnn} training, with customized configurations tailored to our specific application requirements. The trained \textsc{Bnn} models are adopted to predict the EoL of new batteries and provide detailed certainty quantification based on their sensor readings. Additionally, the \textsc{Bnn} models are designed to be adaptive to update the predictions as new sensor data becomes available, improving both the accuracy and certainty of the predictions. 
In summary, we have the following main contributions in this paper.
\begin{itemize}
    \item We propose a system architecture of battery health monitoring with uncertainty-aware EoL prediction based on real-world battery usage settings;
    \item We design and develop the \textsc{Bnn} models customized for battery EoL prediction with both expected EoL and quantifiable prediction certainty;
    \item We conduct an experimental study to show the effectiveness of the proposed \textsc{Bnn} models, achieving on average 13.9\%, and as low as 2.9\% prediction error rate with 66\% certainty improvement from cycle 100 to 400.
\end{itemize}


The rest of the paper is organized as follows. We present our system architecture of \textsc{Bnn}-based battery EoL prediction in Section \ref{sec:sys}. We describe our methodology of data and technical details of \textsc{Bnn} in Section \ref{sec:method}. In Section \ref{sec:exp}, we present the experimental study and results analysis. Finally, we conclude this paper in Section \ref{sec:conclusion}.

\section{System Architecture and Description}
\label{sec:sys}
In this section, we present the system architecture of battery EoL prediction using the uncertainty-aware \textsc{Bnn}.

\subsection{Overview of A Battery Health Monitoring System}
\begin{figure}
    \centering	\includegraphics[width=1\linewidth]{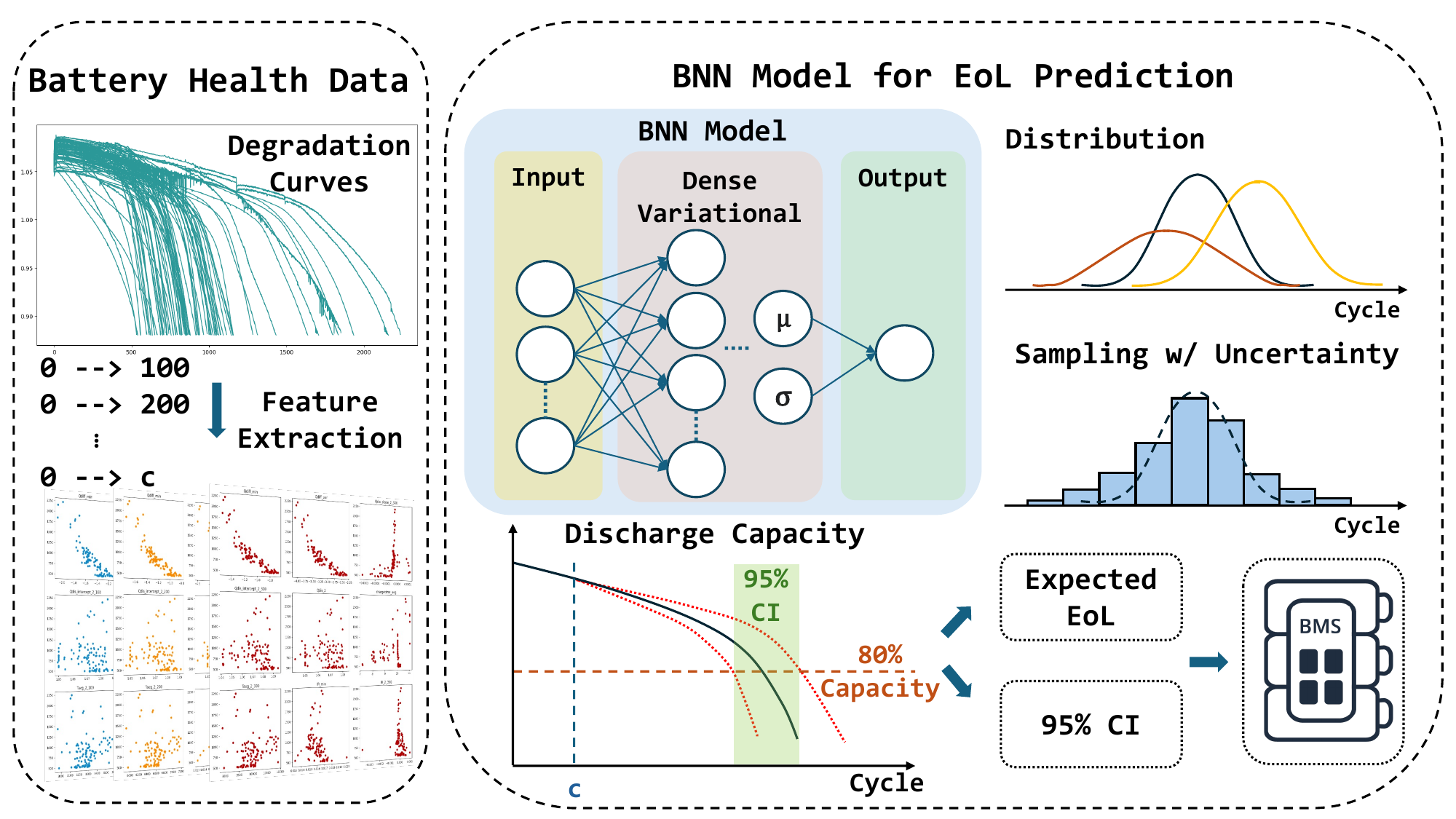}
    \caption{An illustration of the system architecture of ML-based battery EoL prediction. Using the battery's health data about charging and discharging, an ML model predicts the battery's expected EoL together with a quantifiable certainty as a 95\% confidence interval (CI). The prediction results are used by the BMS to offer early EoL alerts and other health-related services.}
    \label{fig:sys}
\end{figure}

Battery health monitoring is typically part of a battery management system (BMS), which uses various sensors to track the battery's dynamics during charge and discharge throughout its lifetime. Some sensor readings are used directly for health monitoring while some others require data transformation and feature extraction to show a stronger correlation with battery health. Identifying a set of features that optimally correlates with battery health is a significant area of research, though not the focus of this paper. Once suitable features are identified, various methods, ML- and non-ML-based, can be utilized to access or predict the battery's health and its remaining useful life. The prediction results are then communicated to the BMS, for subsequent battery usage and related services. An illustration of the system is presented in Fig. \ref{fig:sys}.

\subsection{Battery Health Indicators}
Battery health can be accessed through various measures and many focus on the battery's maximum capacity. During a battery's usage, the state-of-charge (SoC) indicates the current charge level, as a percentage calculated as $(Q^{\text{current}}/Q^{\max})\times 100\%$, where $Q^{\text{current}}$ and $Q^{\max}$ represent the battery's current and maximum capacity, respectively. Typically, $Q^{\max}$ and 100\% SoC are achieved when a new battery is fully charged. Due to battery aging, the maximum achievable SoC declines with increased charging/discharging cycles. In EVs, a battery is considered to have reached its EoL when its SoC fails to exceed 80\% following a full charge, marked as cycle $c^\text{EoL}$. This paper aims to predict such $c^\text{EoL}$, interchangeable with EoL in this paper for simplicity.

\subsection{Uncertainty-Aware EoL Prediction}
Most ML models for EoL prediction typically generate a single predicted value for each tested battery. This value indicates the expected EoL, i.e., $\mathbb{E}(c^\text{EoL})$, at which the EoL is the most likely to happen. However, relying solely on expected EoL can be insufficient in practical applications and sometimes even misleading. In addition to $\mathbb{E}(c^\text{EoL})$, it would be useful for the BMS to understand that there is high confidence that the battery will not reach EoL before $\mathbb{E}(c^\text{EoL}) - \Delta c$, where $0<\Delta c<\mathbb{E}(c^\text{EoL})$ is an integer. Beyond $\mathbb{E}(c^\text{EoL}) - \Delta c$, battery health should become a concern, as the probability of reaching EoL increases significantly, even if it is yet $\mathbb{E}(c^\text{EoL})$. Otherwise, there is a risk of unexpected battery failures, which can be particularly hazardous during operations like driving. 

Besides, data scarcity underscores the need for certainty-aware EoL prediction. The limited availability of battery health data presents a daunting challenge for many battery-related research efforts. The scarcity is often due to the expensive data collection for both cost and time, as well as the limited diversity of battery data across different usage settings and environments. For example, the well-known Stanford-MIT dataset \cite{severson2019data}, which is used in this study, consists of only about a hundred battery aging curves. Such a limited dataset is far from enough for training an ML model that can deliver predictions with high certainty and confidence. Such limitation should be reflected in battery monitoring, as $\Delta c$ could be too significant to overlook. Thus in this paper, we aim to develop ML models for EoL predictions that are uncertainty-aware and risk-minimized. Specifically, we propose to use \textsc{Bnn} to address the challenges.

\subsection{\textsc{Bnn} for EoL Prediction}
In this paper, we choose \textsc{Bnn} for ML-based EoL prediction due to \textsc{Bnn}'s key advantage of capturing prediction uncertainties. Like most ML models, a \textsc{Bnn} model consists of a set of parameters. However, unlike traditional ML models that treat parameters as fixed values, \textsc{Bnn} models each parameter as a distribution. This approach allows the parameter values to vary according to the probability distributions with varying predictions even for the same input. This variability is the main feature of \textsc{Bnn} and brings it the capability of modelling uncertainty effectively by extracting statistical features from a list of predictions. We defer the detailed technical description of \textsc{Bnn} to Section \ref{sec:method}. Our main point here is that \textsc{Bnn} not only predicts the expected EoL but also provides quantifiable confidence, e.g., 95\% CI, that is essential for practical battery health monitoring. Overall, the description of the system architecture paves the way for our detailed technical description in the following methodology section.

\section{Methodologies}
\label{sec:method}
Our methodology begins with an overview of the data used in this research, followed by the technical details of the \textsc{Bnn}.

\subsection{Battery Health Monitoring Data and Features}
Here, we present the dataset and the features for \textsc{Bnn}.

\subsubsection{Dataset}
We focus on Li-ion batteries and use the popular Stanford-MIT dataset \cite{severson2019data}. This dataset comprises 124 commercial cells, each with a 1.1 Ah nominal capacity, tested in a temperature-controlled environmental chamber. Though the dataset is not considered as large by ML standards, it provides several key advantages for battery health monitoring. These include a wide range of EoL from 150 to 2,300 cycles, different charging policies, and comprehensive recordings of internal impedance, voltage, current, cell temperature, etc. Such diversity in data is crucial for our ML analysis.

\subsubsection{Features}
Besides the dataset, \cite{severson2019data} also identified some features that significantly correlate with battery health and EoL. The paper evaluated three ML models, each utilizing a different set of features, and found that the full model incorporating all the features performs the best. In our research, we follow this finding and use all features in \textsc{Bnn}. We group them into three categories detailed in Table \ref{tab:features}. 

\begin{table}
\caption{The features used in \textsc{Bnn} for EoL estimation at cyle $c$. Three feature categories are for $\Delta Q$, discharge, and others.}
\centering
\renewcommand{\arraystretch}{1.3}
\begin{tabular}{c|c|c}
\hline\hline
Category & Cycles & Features \\ \hline
\multirow{2}{*}{\parbox{6em}{\centering $\Delta Q_{c-10}(V)$ curve}} & $10,c$ & minimum $\Delta Q$ in the curve \\
& $10,c$ & variance $\Delta Q$ in the curve \\ \hline

\multirow{3}{*}{\parbox{6em}{\centering discharge capacity fade curve}} & $2,\ldots,c$ & slope of the curve's linear regression \\
& $2,\ldots,c$ & intercept of the curve's linear regression \\ 
& $2$ & discharge capacity \\ \hline

\multirow{4}{*}{\parbox{6em}{\centering others}} & $1,\ldots,5$ & average charge time \\
& $2,\ldots,c$ & integral of temperature over time \\ 
& $2,\ldots,c$ & minimum internal resistance \\ 
& $2,c$ & difference of internal resistance \\ \hline
\hline
\end{tabular}
\label{tab:features}
\end{table}

The first category is based on the empirical knowledge that a battery's discharge voltage curve $Q(V)$ tends to flatten with increased battery charge/discharge cycles. So, the difference between the two curves from different cycles becomes a critical indicator for estimating the battery's degradation. We define this difference as $\Delta Q_{c-c'}(V)$ for cycles $c$ and $c'$ where $c>c'$. In our study, we use \textsc{Bnn} to monitor EoL at specific cycles, e.g., $c=200$, with $c'$ fixed to 10 for simplicity. This approach allows us to represent the curve as $\Delta Q_{c-10}$ and we extract the minimum and variance (subject to interpolation) of the curve as the first two features. 

The features in the remaining two categories are relatively easier to obtain. One category involves the discharge capacity fade curve, derived from the maximum capacity after a full charge. Intuitively the curve declines eventually when a battery ages. To simplify, we use a linear regression to approximate this curve. From the regression line, we extract the slope and intercept as features, and the capacity of the second cycle is used as a reference point. The last category includes five features that are relatively independent of each other. One is the average charge time over the first five cycles. The rest features consider all the cycles from the second onward, including the integral of temperatures and both the minimum and difference in internal resistance. 

\subsection{\textsc{Bnn} for Uncertainty-Aware EoL Prediction}
The features presented above are used as the input of our \textsc{Bnn} model. \textsc{Bnn} quantifies uncertainty along with its prediction and this feature serves our research objective well for incorporating uncertainty into ML-based EoL prediction. An ML model consists of parameters, which commonly remain unchanged once the model is trained. \textsc{Bnn} has a different concept that a parameter should not be a constant, instead it should vary based on its probability distribution. The technical details of \textsc{Bnn} are presented as follows.

\subsubsection{Training}
During model training, a \textsc{Bnn} model is first initialized with an assumed prior distribution $p(w)$ for each parameter $w$. This assumption expresses the model's initial belief about the parameter, and such belief is often just a randomized state without observing any data. After initialization, dataset $\mathcal{D}=(X, C^\text{EoL})$ with features $X$ and ground-truth $C^\text{EoL}$ is used to optimize the distribution for each parameter. This essentially is a process of searching for posterior distribution $p(w|\mathcal{D})$ of each parameter $w$ by considering $\mathcal{D}$. Such distribution can be computed through Bayes' rule as,
\begin{equation}
\label{eq:pwd}
    p(w|\mathcal{D}) = p(w|X,C^\text{EoL}) = \frac{p(C^\text{EoL}|X,w)\times p(w)}{p(C^\text{EoL}|X)},
\end{equation}
where the two components in the numerator are easy to calculate and the denominator is computing intensive. Specifically, $p(C^\text{EoL}|X,w)$ is the likelihood derived by applying the latest \textsc{Bnn} parameters to the given input $X$, a typical computing-lite ML inference process. $p(w)$ reflects the model's latest belief and is readily available. $p(C^\text{EoL}|X)$ is the marginal likelihood where all the possible parameters need to be considered, i.e.,
\begin{equation}
\label{eq:pyx}
    p(C^\text{EoL}|X) = \int_{w'}p(C^\text{EoL}|X,w')\times p(w')dw',
\end{equation}
where the computing potentially involves high-dimension space. In practice, such computationally demanding task is replaced with approximation methods such as variational inference or Monte-Carlo for efficient training. This however is not our focus and please refer to \cite{izmailov2021bayesian} for technical details.

\subsubsection{Inference - Prediction}
With a trained \textsc{Bnn} model, we make a prediction of a new battery based on the model's posterior distributions with sampled parameter values. Different from typical ML models, a \textsc{Bnn} model has different sets of parameter values and each set allows the model to make one prediction. Overall, we have a list of predictions for the same input and aggregate the predictions to generate the final output, e.g., statistical information about the predictions. In this paper, we present the aggregated results as a distribution, from which we can quantify the prediction uncertainty with expected EoL. Often, such results follow a Gaussian distribution based on the central limit theorem. Given a list of $n$ predictions $\hat{\mathbf{c}}^\text{EoL}=\hat{c}^\text{EoL}_1, \ldots, \hat{c}^\text{EoL}_n$, both mean $\mu$ and standard deviation (SD) $\sigma$ can be calculated to fit a Gaussian distribution $\mathcal{N}(\hat{\mathbf{c}}^\text{EoL}; \mu, \sigma^2)$,
with $\mu = \frac{1}{n} \sum_{i=1}^n \hat{c}^\text{EoL}_i$ for expected EoL and $\sigma = \sqrt{\frac{1}{n-1} \sum_{i=1}^n (\hat{c}^\text{EoL}_i - \mu)^2}$ for uncertainty.

\section{Experimental Study}
\label{sec:exp}
We evaluate and discuss \textsc{Bnn}'s EoL prediction performance in this section and let us present our experimental setup first.

\subsection{Experimental Setup}
This part includes \textsc{Bnn}'s model configuration, training strategy, and other information like computing resources. 

\subsubsection{Model Configuration}
\textsc{Bnn}'s model configuration is an important part of our experimental study. The model begins with an input layer, which takes in the features we extracted from battery data. The layer is connected with multiple \texttt{DenseFlipout} layers, a fully connected (FC) layer with 2 neurons, and an \texttt{IndependentNormal} sampling layer. In each \texttt{DenseFlipout} layer, there is a \texttt{Flipout} estimator \cite{wen2018flipout} for performing a Monte-Carlo approximation of \textsc{Bnn}'s posterior distribution for variational inference. The estimator can achieve lower variance for kernel and bias compared to other estimators. The \texttt{IndependentNormal} is configured by assuming Gaussian distributions of the final prediction and naturally the layer samples the predicted EoL following the distribution information of both mean and SD. Note that there is an FC layer right before the \texttt{IndependentNormal} layer. This is mainly for mapping the output features from the last \texttt{DenseFlipout} layer to the \texttt{IndependentNormal} layer as specific values corresponding to the mean and SD of EoL.


\subsubsection{Training Strategy} 
Our \textsc{Bnn} searches for its optimal posterior parameter distributions with a multi-stage training strategy. We set a relatively large initial learning rate of 0.05, for coarse-grained fast search. When the training performance, measured by mean absolute error (MAE), has not improved for 10 epochs, we reduce the learning rate by half and trigger a new training stage with a finer-granularity search. We perform multiple stages with a decreased learning rate until it reaches 0.001. Furthermore, the training can stop earlier given the model performance remains non-improved for 30 epochs. 

\subsubsection{Others}
For results analysis, we sample the \textsc{Bnn} output 100 times and estimate the EoL distribution of the tested battery with mean and SD. All experiments are run on a workstation with an AMD Ryzen 9 5950X processor and NVIDIA GTX 3080 GPU. The reported experiment results are based on 10 independent runs by default for cross-validation, each run with randomly selected 80\% batteries for training and 20\% for testing. Some batteries from the dataset reached EoL in the early stages ($c^\text{EoL}< 500$) and are excluded.

\begin{figure}
    \centering	\includegraphics[width=0.98\linewidth]{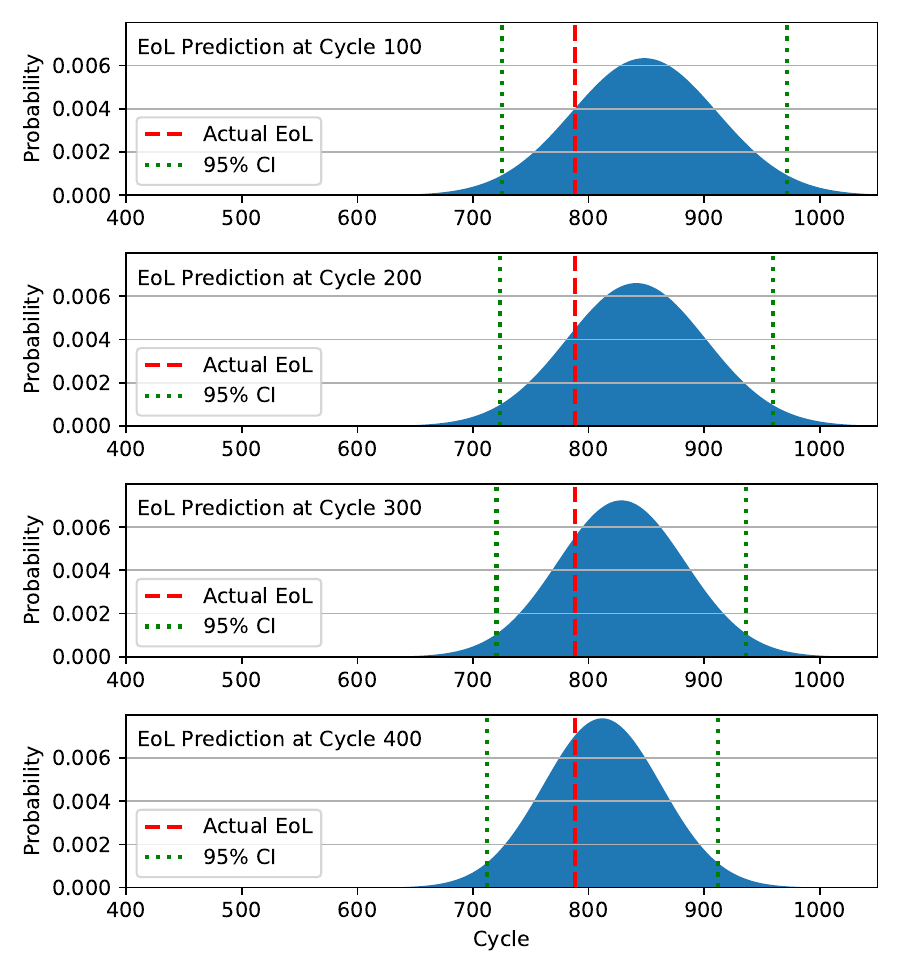}
    \caption{A case study of \textsc{Bnn}'s performance on a specific battery cell showing prediction distribution at cycles 100, 200, 300, and 400. In each sub-figure, the battery's actual EoL is shown as a dashed vertical line at cycle 788. Both the lower limit and upper limit of the 95\% CI in each sub-figure are shown in dotted vertical lines. \textsc{Bnn}'s EoL prediction achieves improved accuracy and certainty with more cycles of information.}
    \label{fig:case_study}
\end{figure}

\subsection{\textsc{Bnn} Case Study for EoL Prediction}
A case study is the most direct and intuitive way to show \textsc{Bnn}'s performance for EoL prediction. Here, we show the results of one tested battery in Fig. \ref{fig:case_study}. There are four sub-figures that show the EoL prediction results at cycles 100, 200, 300, and 400, respectively. The \textsc{Bnn} model's prediction in each sub-figure is reflected as a distribution statistically derived from the model's prediction samples. The mean is the cycle that corresponds to the peak of the distribution and SD measures the spread. A well-performed model should produce a mean EoL that is close to the actual EoL and a small SD which indicates high certainty of the prediction. Besides, we quantify the certainty with 95\% CI, which is calculated as $[\mu - z\times\sigma, \mu + z\times\sigma]$, where $z=1.96$ as the $z$-score. 

\subsubsection{Expected EoL}
Seen from Fig. \ref{fig:case_study}, \textsc{Bnn} estimates the EoL well where the peak of the distribution is not far from the actual EoL. \textsc{Bnn} is optimistic about the tested battery, with the distribution peak occurring tens of cycles after the actual EoL. As the battery approaches its EoL, the prediction accuracy improves. For prediction at cycle 100, the gap between the expected EoL and ground truth is 60.4 cycles. With more updated battery data at cycle 200, the gap reduces to 53.3 with 11.8\% improvement. The gap continues to shrink to 40.5 for cycle 300 and eventually to 23.9 for cycle 400. Such a minimal gap implies 60.4\% of the prediction error reduction compared to cycle 100 and shows \textsc{Bnn}'s competitive performance for this battery with merely 3\% prediction error.

\subsubsection{Prediction Uncertainty}
\textsc{Bnn} also quantifies the prediction certainty. For the prediction at cycle 100, \textsc{Bnn} predicts that battery health would not be an issue before cycle 725 and the EoL can be up to 972 with significant likelihood based on the 95\% CI. Indeed, the actual EoL lies within the interval. Such certainty-aware prediction is important to trigger a non-delayed warning indicating that the battery may reach its EoL earlier by up to 123 cycles with non-negligible probability. This is especially important when the model is optimistic about the battery health. Because it is a huge risk to assume a normal working condition of the battery until its expected EoL of 848, while the battery reaches EoL 60 cycles before at cycle 788. ML's EoL prediction accuracy is likely to be better in the future, but a full certainty of its prediction may never be achieved. Quantifying and understanding the limits of an ML model is among the prerequisites of deploying it for battery health monitoring in practice with confidence.

\subsubsection{Uncertainty Trend}
A trend of the \textsc{Bnn}'s uncertainty can be observed where \textsc{Bnn} shows increased confidence about its prediction when it is made in the later stage of battery usage. The CI as shown in the sub-figures continues to narrow from $\sim$250 for cycle 100 prediction to below 200 with 400 cycles information, with a sharper distribution. The peak probability of the distribution increases from about 0.6\% to about 0.8\% which is over 30\% higher. Such a trend is encouraging as it matches well with real-world battery usage. Typically, the battery users are not too concerned in the beginning stage and become more sensitive to the prediction certainty when the battery is close to its EoL. Our \textsc{Bnn} models can produce more and more certain predictions over time.

\begin{table}
\caption{\textsc{Bnn}'s overall performance at different prediction cycles. We report MAE and MAPE of $\mu$ and $\sigma$ for certainty.}
\centering
\renewcommand{\arraystretch}{1.3}
\begin{tabular}{c|cc|cc|cc}
\hline\hline
\multirow{2}{*}{\begin{tabular}[c]{@{}c@{}}Prediction\\ Cycle\end{tabular}} & \multicolumn{2}{c|}{MAE} & \multicolumn{2}{c|}{MAPE (\%)} & \multicolumn{2}{c}{Uncertainty $\sigma$} \\ \cline{2-7} 
& Train & Test & Train & Test & Train & Test \\ \hline
100 & 160.8 & 183.5          & 17.3 & 18.3           & 60.4 & 66.5           \\ \hline
200 & 154.7 & 160.0          & 15.4 & 16.4           & 52.1 & 61.1           \\ \hline
300 & 113.1 & 131.2       & 11.4 & 14.6           & 49.2 & 58.3           \\ \hline
400 & 84.2  & \textbf{105.6} & 10.4 & \textbf{13.9}  & 34.6 & \textbf{40.1}  \\ \hline
\hline
\end{tabular}
\label{tab:stat}
\end{table}

\subsection{Comprehensive Performance Analysis}
Besides the case study of one tested battery, we report the prediction performance for all batteries and show the results in Table \ref{tab:stat}. We report the statistical performance in terms of MAE and mean absolute percentage error (MAPE) for both training and testing. A unique advantage of \textsc{Bnn} is certainty quantification and we include $\sigma$ as well as a certainty indicator. 

\subsubsection{Trends based on Statistical Information}
The results reveal similar trends as we discussed in the case study. The MAE measures the distance between the expected EoL and the actual EoL. It is over 180 for the prediction at cycle 100. Then it becomes smaller with more cycling information available, e.g., 160 for cycle 200 and eventually to a bit over 100 with 400 cycles. MAPE describes errors as a percentage of the actual EoL and its results show the same trend. The percentage is relatively high at 18.3\% at cycle 100 and improves to 13.9\% at cycle 400 with a 24\% error reduction. \textsc{Bnn} not only is more accurate but also reduces uncertainty with more cycling data, with $\sigma$ reducing from 66.5 for cycle 100 to 40.1 for cycle 400 where the initial uncertainty is 66\% higher. Overall, \textsc{Bnn} offers both accurate and certainty-aware EoL predictions which improve throughout a battery's lifetime. 

\subsubsection{Over-fitting}
Another observation is that \textsc{Bnn} models can be over-fitted. In most runs of experiments, the prediction performance is more competitive during training compared to testing. At cycle 400, \textsc{Bnn}'s prediction error is 84.2 for the batteries used for training and 105.6 when new batteries are tested, with a 25\% higher error. The prediction uncertainty $\sigma$ also increases from 34.6 to 40.1 by 16\%. Various \textsc{Bnn} model configurations have been tested and such a performance gap between training and testing remains true in different settings. One issue is the dataset size, just over a hundred. The scale of the data fails to offer sufficient diversity for an ML model to learn to be generic. As a result, a model's optimized parameters for training data cannot fully generalize to the testing data. We expect that the over-fitting issue will be alleviated with more data available in the future.

\begin{figure}
    \centering	
    \def \tmpw{0.5}
    \subfigure[]{\includegraphics[width=\tmpw\linewidth]{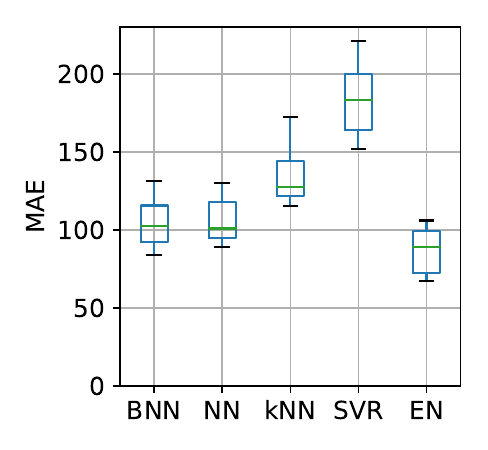}\label{fig:CompMAE}}
    \hspace{-1em}
    \subfigure[]{\includegraphics[width=\tmpw\linewidth]{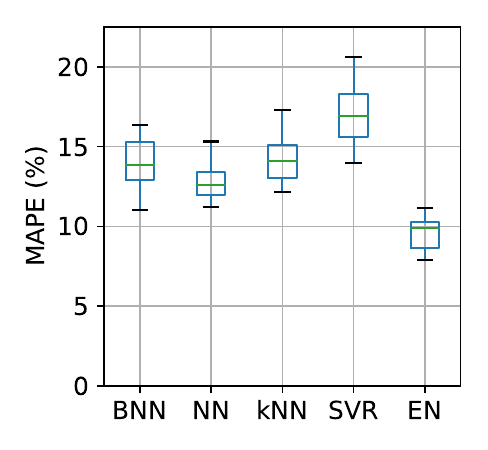}\label{fig:CompMAPE}}
    \caption{A comparison study of \textsc{Bnn} and its counterparts with the same data but different ML models for MAE in Fig. \ref{fig:CompMAE} and MAPE in Fig. \ref{fig:CompMAPE}. \textsc{Bnn} achieves competitive prediction accuracy besides its quantifiable uncertainty.}
    \label{fig:comp}
\end{figure}

\subsection{Comparison Study}
\textsc{Bnn} offers uncertainty quantification besides accuracy-based prediction common in most ML models. It is a natural question if such additional benefit is at the cost of other aspects. Here, we compare the MAE and MAPE between \textsc{Bnn} and other ML models, with some of them presented in Fig. \ref{fig:comp}. 

\subsubsection{Models}
\textsc{Bnn} is one of the NN families, and we consider a standard NN as the first ML algorithm for comparison. Two other algorithms are $k$-nearest neighbors ($k$NN) and support vector regression (SVR), widely used in other studies. The last one is elastic net (EN) \cite{zou2005regularization}, demonstrated to be effective in \cite{severson2019data}. The optimal model configurations are searched for all comparison models and the best configurations are chosen for this study. Note that none of the comparison models is capable of incorporating certainty in predictions.

\subsubsection{Results}
Seen from Fig. \ref{fig:CompMAE}, \textsc{Bnn}, despite its additional uncertainty awareness, achieves similar prediction accuracy to its counterparts. For MAE, \textsc{Bnn} shares the same level of performance with NN with both achieving an average error of about 100 cycles. $k$NN and SVR are traditional ML models proposed decades ago, with an average mis-prediction of 127 and 183 cycles, respectively. \textsc{Bnn} outperforms both. EN has been shown to be suitable for battery health analysis and is about ten cycles more accurate than \textsc{Bnn}. Worth mentioning that EN however does not capture uncertainty in its predictions. This can be one factor that EN is not adopted in certain applications where informing users of prediction certainty is of vital importance. 

The MAPE results in Fig. \ref{fig:CompMAPE} show similar observations, e.g., \textsc{Bnn} performs similarly to NN and $k$NN and outperforms SVR. However, the performance gap between different models varies compared to Fig. \ref{fig:CompMAE}, e.g., $k$NN is not worse than \textsc{Bnn} in terms of MAPE. This is mainly because of the inherent randomness of our experiments. A big absolute error does not always imply a big percentage error, as the actual EoL can be large for the tested battery. On the contrary, when the actual EoL is small, even a minor mis-prediction may change the percentage significantly. This is one of the practical considerations of real-world deployment of battery health monitoring solutions. A more suitable performance evaluation metric that better serves the application shall be selected.

\section{Conclusion}
\label{sec:conclusion}
In this paper, we presented a practical battery health monitoring system with uncertainty-aware EoL prediction using \textsc{Bnn}. We recognized the importance of quantifying uncertainty for battery health monitoring with practical benefits such as safety and sustainability. We designed \textsc{Bnn}-based solutions for analyzing the battery's sensor readings and extracted features to predict the battery's expected EoL with uncertainty quantification. We conducted an experimental study and demonstrated \textsc{Bnn}'s important roles of integrating certainty with EoL prediction with a case study and statistical analysis. Our \textsc{Bnn} models achieved an average prediction error rate of 13.9\%, and the rate can be as low as 2.9\% for certain tested batteries. We also observed the trends that \textsc{Bnn} becomes more accurate with improved certainty when more cycling information is available, and the certainty improvement is 66\% for the predictions from cycle 100 to 400. Moreover, \textsc{Bnn}'s prediction accuracy is competitive compared to several popular ML algorithms, despite \textsc{Bnn}'s additional benefit of certainty awareness. Overall, \textsc{Bnn} can be an important health monitoring module in the BMS and support various battery health-related services.


\bibliographystyle{IEEEtran}
\bibliography{reference}

\begin{thebibliography}{10}
\providecommand{\url}[1]{#1}
\csname url@samestyle\endcsname
\providecommand{\newblock}{\relax}
\providecommand{\bibinfo}[2]{#2}
\providecommand{\BIBentrySTDinterwordspacing}{\spaceskip=0pt\relax}
\providecommand{\BIBentryALTinterwordstretchfactor}{4}
\providecommand{\BIBentryALTinterwordspacing}{\spaceskip=\fontdimen2\font plus
\BIBentryALTinterwordstretchfactor\fontdimen3\font minus \fontdimen4\font\relax}
\providecommand{\BIBforeignlanguage}[2]{{%
\expandafter\ifx\csname l@#1\endcsname\relax
\typeout{** WARNING: IEEEtran.bst: No hyphenation pattern has been}%
\typeout{** loaded for the language `#1'. Using the pattern for}%
\typeout{** the default language instead.}%
\else
\language=\csname l@#1\endcsname
\fi
#2}}
\providecommand{\BIBdecl}{\relax}
\BIBdecl

\bibitem{zhao2024batsort}
Y.~Zhao, W.~Zhang, E.~Hu \emph{et~al.}, ``Batsort: Enhanced battery classification with transfer learning for battery sorting and recycling,'' \emph{arXiv preprint arXiv:2404.05802}, 2024.

\bibitem{zhao2022lithium}
S.~Zhao, C.~Zhang, and Y.~Wang, ``Lithium-ion battery capacity and remaining useful life prediction using board learning system and long short-term memory neural network,'' \emph{Journal of Energy Storage}, vol.~52, p. 104901, 2022.

\bibitem{9208399}
Q.~Liu, Y.~Kang, S.~Qu \emph{et~al.}, ``An online soh estimation method based on the fusion of improved ica and lstm,'' in \emph{2020 IEEE/IAS Industrial and Commercial Power System Asia (I\&CPS Asia)}, 2020, pp. 1163--1167.

\bibitem{severson2019data}
K.~A. Severson, P.~M. Attia, N.~Jin \emph{et~al.}, ``Data-driven prediction of battery cycle life before capacity degradation,'' \emph{Nature Energy}, vol.~4, no.~5, pp. 383--391, 2019.

\bibitem{alipour2022improved}
M.~Alipour, S.~S. Tavallaey, A.~M. Andersson \emph{et~al.}, ``Improved battery cycle life prediction using a hybrid data-driven model incorporating linear support vector regression and gaussian,'' \emph{ChemPhysChem}, vol.~23, no.~7, p. e202100829, 2022.

\bibitem{fei2021early}
Z.~Fei, F.~Yang, K.-L. Tsui \emph{et~al.}, ``Early prediction of battery lifetime via a machine learning based framework,'' \emph{Energy}, vol. 225, p. 120205, 2021.

\bibitem{8418374}
L.~Ren, L.~Zhao, S.~Hong \emph{et~al.}, ``Remaining useful life prediction for lithium-ion battery: A deep learning approach,'' \emph{IEEE Access}, vol.~6, pp. 50\,587--50\,598, 2018.

\bibitem{shen2020hybrid}
S.~Shen, V.~Nemani, J.~Liu \emph{et~al.}, ``A hybrid machine learning model for battery cycle life prediction with early cycle data,'' in \emph{2020 IEEE Transportation Electrification Conference \& Expo (ITEC)}.\hskip 1em plus 0.5em minus 0.4em\relax IEEE, 2020, pp. 181--184.

\bibitem{9694435}
Y.~Ke, R.~Zhou, R.~Zhu \emph{et~al.}, ``State of health estimation of lithium ion battery with uncertainty quantification based on bayesian deep learning,'' in \emph{2021 3rd International Conference on System Reliability and Safety Engineering (SRSE)}, 2021, pp. 12--18.

\bibitem{izmailov2021bayesian}
P.~Izmailov, S.~Vikram, M.~D. Hoffman \emph{et~al.}, ``What are bayesian neural network posteriors really like?'' in \emph{International conference on machine learning}.\hskip 1em plus 0.5em minus 0.4em\relax PMLR, 2021, pp. 4629--4640.

\bibitem{wen2018flipout}
Y.~Wen, P.~Vicol, J.~Ba \emph{et~al.}, ``Flipout: Efficient pseudo-independent weight perturbations on mini-batches,'' 2018.

\bibitem{zou2005regularization}
H.~Zou and T.~Hastie, ``Regularization and variable selection via the elastic net,'' \emph{Journal of the Royal Statistical Society Series B: Statistical Methodology}, vol.~67, no.~2, pp. 301--320, 2005.

\end{thebibliography}

\end{document}